\documentclass[
]{ceurart}

\usepackage{minted}
\setminted{breaklines=true}

\begin{document}

\copyrightyear{2025}
\copyrightclause{Copyright for this paper by its authors.
  Use permitted under Creative Commons License Attribution 4.0
  International (CC BY 4.0).}

\conference{De-Factify 4: Fourth Workshop on Multimodal Fact-Checking and Hate Speech Detection, February, 2025, Philadelphia, USA}

\title{NAU-QMUL: Utilizing BERT and CLIP for Multi-modal AI-Generated Image Detection}

\author[1]{Xiaoyu Guo}[%
orcid=0000-0002-0901-2222,
email=270732@nau.edu.cn,
]
\address[1]{School of Accounting, Nanjing Audit University (NAU),
Nanjing, Jiangsu, China}

\author[2]{Arkaitz Zubiaga}[%
orcid=0000-0003-4583-3623,
email=a.zubiaga@qmul.ac.uk,
url=http://www.zubiaga.org/,]
\address[2]{School of Electronic Engineering and Computer Science, Queen Mary University of London (QMUL), London, UK}

\begin{abstract}
  With the aim of detecting AI-generated images and identifying the specific models responsible for their generation, we propose a multi-modal multi-task model. The model leverages pre-trained BERT and CLIP Vision encoders for text and image feature extraction, respectively, and employs cross-modal feature fusion with a tailored multi-task loss function. Additionally, a pseudo-labeling-based data augmentation strategy was utilized to expand the training dataset with high-confidence samples. The model achieved fifth place in both Tasks A and B of the `CT2: AI-Generated Image Detection' competition, with F1 scores of 83.16\% and 48.88\%, respectively. These findings highlight the effectiveness of the proposed architecture and its potential for advancing AI-generated content detection in real-world scenarios. The source code for our method is published on https://github.com/xxxxxxxxy/AIGeneratedImageDetection.

\end{abstract}

\begin{keywords}
  Multi-modal \sep
  AI-generated content detection \sep
  Multi-task \sep
  Pseudo-labeling
\end{keywords}

\maketitle

\section{Introduction}

The rapid evolution of text-to-image generation systems has significantly transformed creative and practical domains, enabling the seamless generation of high-quality images from textual descriptions. Models such as Stable Diffusion \cite{rombach2022high}, DALL-E \cite{ramesh2022hierarchical}, and MidJourney\footnote{https://www.midjourney.com} represent state-of-the-art techniques in this field, capable of producing high-quality images with high levels of detail and realism. However, this technological leap has introduced critical challenges in distinguishing AI-generated images from those created by humans, which is paramount to ensure the authenticity of media, to protect intellectual property, and to combat misinformation \cite{Bontridder_Poullet_2021}.

To address these challenges, we set out to use multi-modal deep learning models to extract and learn the unique features of various image generation systems, enabling robust classification of generated images. By integrating the strengths of models such as BERT \cite{devlin2019bert} for textual analysis and CLIP \cite{radford2021learning} for image representation, our approach integrates cross-modal information to identify distinctive characteristics of each generation model. This methodology enabled our approach to secure the fifth place in both Tasks A and B of the CT2: AI-Generated Image Detection competition, achieving an F1 score of 83.16\% and 48.88\%, respectively. These results highlight the robustness of our approach across different tasks.

\section{Background}
\subsection{CT2: AI-Generated Image Detection}
The CT2: AI-Generated Image Detection competition \cite{roy-2025-defactify-overview-image} aims to determine which types of model-generated images are easier or harder to detect, addressing the challenges of distinguishing AI-generated images from those created by humans. It is divided into two tasks. Task A is a binary classification task, in which participants aim to determine whether a given image was generated by AI or created by a human. Task B builds upon Task A and requires participants to identify the specific AI model responsible for generating each image. This includes distinguishing between models such as Stable Diffusion 3 (SD 3), Stable Diffusion XL (SDXL), Stable Diffusion 2.1 (SD 2.1), DALL-E 3, and Midjourney 6.

\subsection{Related Work on Multi-modal AI-Generated Image Detection}
The rapid advancement of large language models has inspired efforts in the vision community to harness large-scale models trained on both images and text \cite{zhou2022learning,zhang2022tip}. While multi-modal approaches have demonstrated significant success in various applications, their utilization in AI-generated image detection remains relatively underexplored.

Recent studies have begun to investigate the use of pre-trained multi-modal models, primarily variants of CLIP, for detecting AI-generated images. Ojha et al. \cite{ojha2023towards} employed nearest neighbor search and linear probing within the CLIP feature space to generalize detection performance to out-of-distribution (OOD) data. However, this approach required a large dataset of fake and real images for classifier training, which can be resource-intensive and limit scalability. Building on these efforts, Cozzolino et al. \cite{cozzolino2024raising} introduced a lightweight CLIP-based detector but demonstrated that superior performance can be achieved using significantly less data. These studies highlight the remarkable flexibility and effectiveness of CLIP as a foundation model for universal AI-generated image detection, showcasing its potential to handle diverse and challenging scenarios with minimal training data.

Despite advancements, the lack of aligned multi-modal datasets remains a key challenge for robust detection methods, especially in triple-modality settings (text, image, and voice). Huangshan et al. \cite{huang2024ru} address this with a large-scale dataset RU-AI, constructed from Flickr8K, COCO, and Places205, containing 1,475,370 instances with AI-generated duplicates and noise variants for robustness analysis. Experiments show that state-of-the-art models still struggle with accurate and robust classification, highlighting the challenges in multi-modal detection.

\section{System Overview}
\begin{figure}
  \centering
  \includegraphics[width=5in]{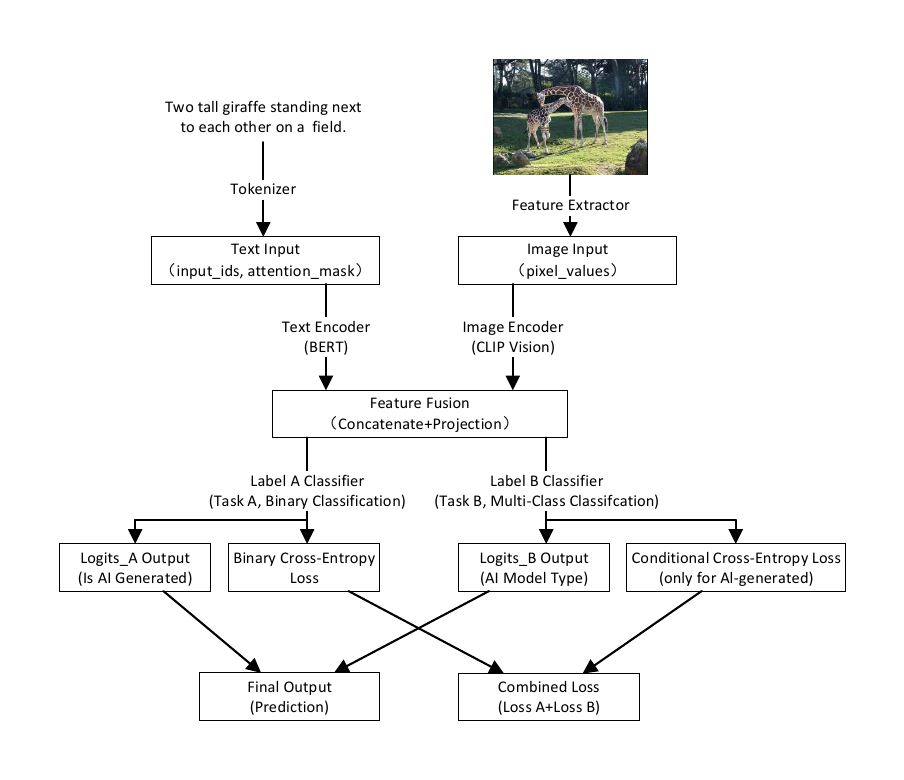}
  \caption{Overall Framework.}
  \label{figure1}
\end{figure}

Figure \ref{figure1} illustrates the architecture of our Multi-modal Multi-task Model, designed to classify and analyze AI-generated images. The model combines text and image inputs through multi-modal feature extraction and fusion, enabling robust predictions for two tasks: detecting whether an image is AI-generated (Task A) and identifying the specific AI model responsible for generating the image (Task B).

\subsection{Feature Extraction and Classification}
The model begins with two input streams: text input and image input. For text input, a tokenizer preprocesses the raw text, converting it into tokenized sequences that are subsequently fed into a pre-trained BERT model to extract contextualized text features. In parallel, the image input is processed into pixel values, which serve as numerical representations of the image, suitable for the model. These pixel values are obtained using a pre-trained AutoFeatureExtractor from the CLIP model. The feature extractor performs standard preprocessing steps, such as resizing the image to a fixed dimension, converting it to a tensor, and normalizing pixel intensity values to align with the requirements of the CLIP vision encoder. This ensures consistency and compatibility across all input images, enabling the vision encoder to extract high-level visual features effectively.

Once the textual and visual features are extracted, they are concatenated and projected into a shared feature space through a fully connected layer, enabling effective multi-modal feature fusion. This shared representation is subsequently passed to two distinct classification heads. The first is the Label\_A Classifier, designed for binary classification to determine whether an image is AI-generated. The second is the Label\_B Classifier, responsible for multi-class classification to identify the specific AI model, such as SD 3, SDXL, DALL-E 3, or MidJourney 6, that generated the image.

\subsection{Multi-task Loss Optimization}
The optimization of this model relies on a multi-task loss function designed for two objectives. For Task A, a Binary Cross-Entropy Loss is applied to classify images as either real or AI-generated. For Task B, a Conditional Cross-Entropy Loss is computed only for samples that Task A has classified as AI-generated (i.e., LABEL\_A == 1). Since LABEL\_B is only meaningful when LABEL\_A predicts an AI-generated image, this conditional design prevents unnecessary computations for real images (LABEL\_A == 0), reducing noise and improving classification efficiency. The total loss is the sum of these two components, ensuring the model effectively balances both tasks.

\section{Experimental Setup}
\subsection{Dataset}
The dataset used for our experiments was released by the organizers of the CT2: AI-Generated Image Detection \cite{roy-2025-defactify-dataset-image}. The dataset comprises 53,353 samples derived from the original MS COCO dataset, where captions and images were processed using various text-to-image generation models, including SD 3, SDXL, SD 2.1, DALL-E 3, and Midjourney 6. Captions from MS COCO were fed into these models to generate the corresponding images, creating a diverse and comprehensive dataset.

\subsection{Parameter Setting}
The model integrates a text encoder initialized with the pre-trained BERT model (bert-base-uncased) and an image encoder using the vision module of the CLIP model (openai/clip-vit-base-patch32). The output heads were configured for binary classification (LABEL\_A) and multi-class classification (LABEL\_B) with six categories, including five AI types and one for real images.

The training process was conducted with the following parameters: a learning rate of
$2 \times 10^{-5}$, weight decay set to 0.01, a batch size of 256 for both training and evaluation, and a total of 8 epochs.

The evaluation was performed at the end of each epoch, and the Weighted-F1 score on Task A was selected as the metric to determine the best model. The training process was configured to load the best-performing model at the conclusion of training. Model checkpoints were saved at the end of each epoch, preserving only the most recent checkpoint to reduce storage space.

\subsection{Data Augmentation}
To enhance the generalization ability of the model and improve its performance on AI-generated image detection tasks, we employed a pseudo-label-based data augmentation strategy. This approach leverages high-confidence predictions from the model on unlabeled test data to expand the training dataset with pseudo-labeled samples.

The process begins by using the trained model to predict labels for the test dataset, which consists of captions and their corresponding images. Captions are tokenized using the pre-trained BERT tokenizer, and images are preprocessed into pixel values using the CLIP feature extractor. The model predicts two outputs: Label\_A, indicating whether an image is AI-generated, and Label\_B, specifying the AI model responsible for generating the image. Alongside these predictions, confidence scores are computed for both labels. We set a confidence threshold of 0.8, and only samples where both predictions exceed this threshold are considered for augmentation.

For the selected high-confidence samples, pseudo-labels are generated and stored, including the text captions, image paths, and the predicted labels for Label\_A and Label\_B. These samples are preprocessed and split into training and validation subsets using an 8:2 ratio. The resulting pseudo-labeled datasets are then concatenated with the original training and validation datasets to form the extended training and validation datasets.

\subsection{Implementation}
The execution of our model was carried out on a high-performance computing server equipped with a 20-core CPU, 80 GB NVIDIA A100 GPU, 120 GB RAM, and 400 GB of available workspace\footnote{https://openbayes.com/}.

The software environment was built using PyTorch 2.4, along with Hugging Face’s Transformers library, which facilitated the use of pre-trained models for both text and image encoders. The training and evaluation processes were managed using the Hugging Face Trainer API, which streamlined optimization, evaluation, and checkpoint management.

\section{Results}
During the training phase, the model achieved robust results on the validation set, with an F1 score of 99.58\% for Task A and an Weighted-F1 score of 85.95\% for Task B (multi-class classification). For Task A, the model achieved an accuracy of 99.24\%, with precision and recall scores of 99.44\% and 99.72\%, respectively, reflecting its ability to reliably distinguish between real and AI-generated images. For Task B, the model achieved an accuracy of 90.04\%, with a precision of 82.56\% and a recall of 90.04\%, demonstrating its ability to identify the specific AI models that generated the content.

On the official test set, evaluated by the competition organizers, the model secured the 5th place in both Tasks A and B. For Task A, the model achieved an F1 score of 83.16\%, consistent with its binary classification capability. For Task B, the model achieved an F1 score of 48.88\%, as reported by the organizers.

\section{Conclusion}
This paper presents a multi-modal multi-task model for detecting and analyzing AI-generated images, addressing the challenges of distinguishing AI-generated content and identifying the specific models responsible for generating it. The proposed architecture effectively integrates pre-trained text and image encoders, employs cross-modal feature fusion, and utilizes a tailored multi-task loss function to handle binary and multi-class classification tasks. The competitive performance on the official test set further validated the model's strength, achieving 5th place in both Task A and Task B, with F1 scores of 83.16\% and 48.88\%, respectively.

However, our pseudo-label-based data augmentation strategy introduces potential biases. A key concern is error propagation, where incorrect pseudo-labels reinforce existing model errors. Additionally, selection bias may arise from high-confidence filtering, as our 0.8 threshold favors easier-to-classify samples, underrepresenting ambiguous cases. Data distribution shift is another challenge, as pseudo-labeled samples from the test set may not align with the original training data, affecting generalization. Moreover, class imbalance can occur if certain AI-generated image styles or models receive high-confidence pseudo-labels more frequently, leading to overrepresentation. Finally, train-test contamination poses a risk, as pseudo-labeling test data may introduce overlap between training and evaluation sets, inflating performance estimates.

Future research will focus on exploring more advanced and effective multi-modal feature fusion strategies, moving beyond the simple concatenation approach used in this study. Techniques such as attention mechanisms, cross-modal transformers, or graph-based methods will be investigated to better capture interactions between textual and visual modalities. Additionally, future work will aim to model the relationships between images associated with the same caption, rather than treating them as independent samples. Furthermore, to mitigate the potential biases introduced by our data augmentation strategy, future studies will incorporate uncertainty-aware training, employ diverse augmentation techniques, and implement strategies to address class imbalance, ensuring a more robust and fair model.

\section{Acknowledgments}
During the preparation of this work, the author(s) used ChatGPT in order to improve writing clarity. After using this tool, the author(s) reviewed and edited the content as needed and take(s) full responsibility for the content of the publication.

\end{document}